\documentclass[11pt,a4paper]{article}
\usepackage[hyperref]{acl2019}
\usepackage{times}
\usepackage{latexsym}
\usepackage{booktabs}
\usepackage{multirow}
\usepackage{makecell}
\usepackage{adjustbox}
\usepackage{tikz-dependency}
\usepackage{tikz-qtree}
\usepackage{tikz-qtree-compat}
\usepackage{subcaption} 
\usepackage{tablefootnote}
\newcommand{\stddev}[2]{\ensuremath{#1_{\color{darkgray}{\pm #2}}}} 
\usepackage{booktabs,amsmath}
\newcommand{\NA}{---}

\newcommand{\single}{\textsc{s-s}}
\newcommand{\twoTask}{\textsc{s-mtl}}
\newcommand{\aux}{\textsc{d-mtl-aux}}
\newcommand{\multi}{\textsc{d-mtl}}
\newcommand{\mtl}{\textsc{mtl}}
\newcommand{\las}{\textsc{las}}
\newcommand{\uas}{\textsc{uas}}
\newcommand{\fone}{\textsc{f1}}
\usepackage{cleveref}
\usepackage{url}
\renewcommand{\vec}[1]{\mathbf{#1}}

\newcommand\mybox[2][]{\tikz[overlay]\node[inner sep = 0,
      draw, solid,
      outer sep = .5ex,
      rounded corners = 2pt,
      fill=orange!40, draw=brown, inner sep=.6ex, anchor=text,#1] {#2};\phantom{#2}}

\aclfinalcopy 
\def\aclpaperid{1093} 

\title{Sequence Labeling Parsing by Learning Across Representations}

\author{Michalina Strzyz \qquad David Vilares \qquad Carlos G\'omez-Rodr\'iguez\\
  Universidade da Coru\~na, CITIC \\
  FASTPARSE Lab, LyS Research Group, Departamento de Computaci\'on \\
  Campus de Elvi\~na, s/n, 15071 A Coru\~na, Spain\\
  {\tt \{michalina.strzyz,david.vilares,carlos.gomez\}@udc.es}}

\date{}

\begin{document}
\maketitle
\begin{abstract}

We use parsing as sequence labeling as a common framework to learn across constituency and dependency syntactic abstractions.
To do so, we cast the problem as multitask learning (\textsc{mtl}).
First, we show that adding a parsing paradigm as an auxiliary loss consistently improves the performance on the other paradigm. Secondly, we explore an \textsc{mtl} sequence labeling model that parses both representations, at almost no cost in terms of performance and speed.
The results across the board show that on average \textsc{mtl} models with auxiliary losses for constituency parsing outperform single-task ones by 1.14
\textsc{F1} points, and for
dependency parsing by 0.62 
\textsc{uas} points.\footnote{This is a revision of \url{https://arxiv.org/abs/1907.01339v2}. The previous version contained a bug where the EVALB scripts were not considering the COLLINS.prm and spmrl.prm parameter files.}
  
\end{abstract}

\section{Introduction}

Constituency \cite{chomsky1956three} and dependency grammars \cite{mel1988dependency,kubler2009dependency} are the two main abstractions for representing the syntactic structure of a given sentence, and each of them has its own particularities \citep{depcons}. While in constituency parsing the 
structure of sentences is
abstracted as a phrase-structure tree (see Figure \ref{fig:consTree}), in dependency parsing the tree encodes binary syntactic relations between pairs of words (see Figure \ref{fig:dependecyTree}).

When it comes to developing natural language processing (\textsc{nlp}) parsers, these two tasks are usually considered as disjoint tasks, and their improvements therefore have been obtained separately \cite{charniak2000maximum,nivre2003efficient,kiperwasser2016simple,dozat2016deep,ma2018stack,kitaev2018constituency}. 

Despite the potential benefits of learning across representations, there have been few attempts in the literature to do this.
\newcite{klein2003fast} considered a factored model that provides separate methods for phrase-structure and lexical dependency trees and combined them to obtain optimal parses. With a similar aim, \newcite{ren2013combine} first compute the \emph{n} best constituency trees using a probabilistic context-free grammar, convert those into dependency trees using a dependency model, compute a probability score for each of them, and finally rerank the most plausible trees based on both scores. 
However, these methods are complex
and intended for statistical parsers.
Instead, we propose a extremely simple framework to learn across constituency and dependency representations.

\paragraph{Contribution} (i) We use sequence labeling for constituency \cite{GomVilEMNLP2018} and dependency parsing \citep{viable} combined with multi-task learning (\textsc{mtl}) \cite{caruana} to learn across syntactic representations. To do so, we take a parsing paradigm (constituency or dependency parsing) as an auxiliary task to help train a model for the other parsing representation, a simple technique that translates into consistent improvements across the board. (ii) We also show that a single \textsc{mtl} model following this strategy can robustly produce both constituency and dependency trees, obtaining a performance and speed comparable with previous sequence labeling models for (either) constituency or dependency parsing. The source code is available at \url{https://github.com/mstrise/seq2label-crossrep}

\section{Parsing as Sequence Labeling}

\paragraph{Notation} We use $w=[w_i,...,w_{|w|}]$ to denote an input sentence. We use bold style lower-cased and math style upper-cased characters to refer to vectors and matrices (e.g. $\vec{\bf x}$ and $\vec{W}$).\\ 

\noindent Sequence labeling
is a structured prediction task where each token in the input sentence is mapped to a label \cite{rei2018zero}. Many \textsc{nlp} tasks suit this setup, including part-of-speech tagging, named-entity recognition or chunking \cite{sang2000introduction,toutanova2000enriching,sang2003introduction}. More recently, syntactic tasks such as constituency parsing and dependency parsing have been successfully reduced to sequence labeling \cite{Spoustova,li-EtAl:2018:C18-13,GomVilEMNLP2018,viable}. Such models compute a tree representation of an input sentence using $|w|$ tagging actions.

We will also cast parsing as sequence labeling, to then learn across representations using multi-task learning. Two are the main advantages of this approach: (i) it does not require an explicit parsing algorithm nor explicit parsing structures, and (ii) it massively simplifies joint syntactic modeling.
We now describe parsing as sequence labeling and the architecture used in this work.

\paragraph{Constituency parsing as tagging}

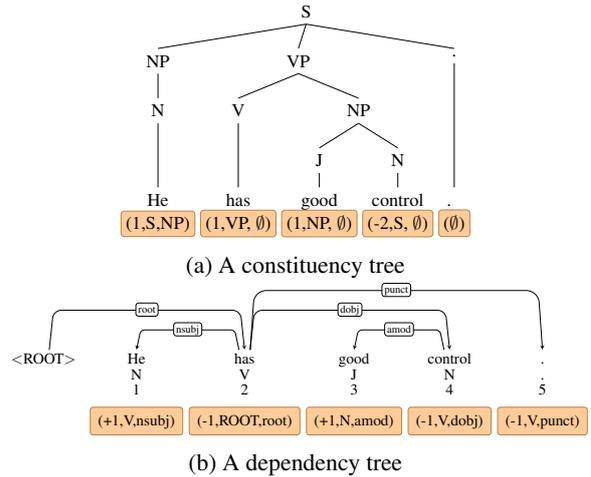
\begin{figure}
\centering
\begin{subfigure}{.6\columnwidth}

\centering
\begin{adjustbox}{max width=\columnwidth}
\begin{tikzpicture}[frontier/.style={distance from root=4cm}]
\tikzset{every tree node/.style={align=center,anchor=north}}
\Tree [.S [.NP [.N He\\\mybox{(1,S,NP)} ]] 
          [.VP [.V has\\\mybox{(1,VP, $\emptyset$)} ] 
              [.NP [.J good\\\mybox{(1,NP, $\emptyset$)} ] [.N control\\\mybox{(-2,S, $\emptyset$)} ]]]
          [. . .\textcolor{white}{P}\\\mybox{($\emptyset$)} ]]]
            
\end{tikzpicture}
\end{adjustbox}
\caption{A constituency tree}
\label{fig:consTree}
\end{subfigure}

\begin{subfigure}{\columnwidth}
\centering
\begin{adjustbox}{max width=\columnwidth}
\begin{dependency}[label theme = default]
   \begin{deptext}[column sep=1em] 
     $<$ROOT$>$ \&  He \&  has \&  good \&  control \& .\\
       \& N \& V \& J \& N \& . \\
        \& 1 \& 2 \& 3 \& 4 \& 5 \\ 
        \\
       \& (+1,V,nsubj) \& (-1,ROOT,root) \& (+1,N,amod) \& (-1,V,dobj) \& (-1,V,punct) \\
   \end{deptext}
   \depedge{3}{2}{nsubj}
    \depedge{3}{5}{dobj}
    \depedge{5}{4}{amod}
     \depedge{1}{3}{root}
     \depedge{3}{6}{punct}
     \wordgroup[group style={fill=orange!40, draw=brown, inner sep=.6ex}]{5}{2}{2}{a0}
     \wordgroup[group style={fill=orange!40, draw=brown, inner sep=.6ex}]{5}{3}{3}{a1}
     \wordgroup[group style={fill=orange!40, draw=brown, inner sep=.6ex}]{5}{4}{4}{a2}
     \wordgroup[group style={fill=orange!40, draw=brown, inner sep=.6ex}]{5}{5}{5}{a3}
     \wordgroup[group style={fill=orange!40, draw=brown, inner sep=.6ex}]{5}{6}{6}{a4}
    
\end{dependency}
\end{adjustbox}
\caption{A dependency tree }
\label{fig:dependecyTree}
\end{subfigure}

\caption{An example of constituency and dependency trees with their encodings.}
\label{fig:trees}
\end{figure}

\newcite{GomVilEMNLP2018} define a linearization method $\Phi_{|w|}: T_{c,|w|} \rightarrow L_{c}^{|w|}$ to transform a phrase-structure tree into a discrete sequence of labels of the same length as the input sentence. Each label $l_i \in L_{c}$ is a three tuple $(n_i, c_i, u_i)$ where: $n_i$ is an integer that encodes the number of ancestors in the tree shared between a word $w_{i}$ and its next one $w_{i+1}$ (computed as  relative variation with respect to $n_{i-1}$), $c_i$ is the non-terminal symbol shared at the lowest level in common between said pair of words, and $u_i$ (optional) is a leaf unary chain that 
connects
$c_i$ to $w_i$. Figure \ref{fig:consTree} illustrates the encoding with an example.\footnote{In this work we do not use the dual encoding by \newcite{VilaresMTL2019}, which combines the relative encoding with a top-down absolute scale to represent certain relations.}

\paragraph{Dependency parsing as tagging} \citet{viable} also propose a linearization method $\Pi_{|w|}: T_{d,|w|} \rightarrow L_{d}^{|w|}$ to transform a dependency tree into a discrete sequence of labels. Each label $r_i \in L_{d}$ is also represented as a three tuple $(o_i,p_i,d_i)$. If $o_i>0$, $w_i$'s head is the $o_i$th closest word with PoS tag $p_i$ to the right of $w_i$. If $o_i < 0$, the head is the $-o_i$th closest word to the left of $w_i$ that has as a PoS tag $p_i$. The element $d_i$ represents the syntactic relation between the head and the dependent terms. Figure \ref{fig:dependecyTree} depictures it with an example.

\paragraph{Tagging with \textsc{lstm}s}

We use bidirectional \textsc{lstm}s (\textsc{bilstm}s) to train our models \cite{hochreiter1997long,schuster1997bidirectional}. Briefly, let $\textsc{lstm}_\rightarrow(\vec{x})$ be an abstraction of a $\textsc{lstm}$ that processes the input from left to right, and let $\textsc{lstm}_\leftarrow(\vec{x})$ be another $\textsc{lstm}$ processing the input in the opposite direction, the output $h_i$ of a $\textsc{bilstm}$ at a timestep $i$ is computed as: $\textsc{bilstm}(\vec{x},i) = \textsc{lstm}_\rightarrow(\vec{x}_{0:i}) \circ \textsc{lstm}_\leftarrow(\vec{x}_{i:|w|})$. Then, $h_i$ is further processed by a feed-forward layer to compute the output label, i.e. $P(y|\vec{h}_i)=\mathit{softmax}(\vec{W}*\vec{h}_i+\vec{b})$. To optimize the model, we minimize the categorical cross-entropy loss, i.e. $\mathcal{L} = -\sum{log(P(y|\vec{h}_i))}$. In Appendix \ref{appendix-training-configuration} we detail additional hyperpameters of the network. In this work we use NCRFpp \cite{yang2018ncrf++} as our sequence labeling framework.

\section{Learning across representations}

To learn across representations we cast the problem as multi-task learning. \textsc{mtl} enables learning many tasks jointly, encapsulating them in a single model and leveraging their shared representation \citep{caruana,ruder}.
In particular, we will use a hard-sharing architecture: the sentence is first processed by stacked \textsc{bilstm}s shared across all tasks, with a task-dependent feed-forward network on the top of it, to compute each task's outputs. In particular, to benefit from a specific parsing abstraction we will be using the concept of auxiliary tasks \cite{plank2016multilingual,benefitsMultitask,aux}, where tasks are learned together with the main task in the \textsc{mtl} setup even if they are not of actual interest by themselves, as they might help to find out hidden patterns in the data and lead to better generalization of the model.\footnote{Auxiliary losses are usually given less importance during the training process.} For instance, \citet{MultitaskParsingSemanticRepresentation} have shown that semantic parsing benefits from that approach.

The input is the same for both types of parsing and the same number of timesteps are required to compute a tree (equal to the length of the sentence), which simplifies the joint modeling. In this work, we focus on parallel data (we train on the same sentences labeled for both constituency and dependency abstractions). In the future, we plan to explore the idea of exploiting joint training over disjoint treebanks \cite{barrett2018sequence}.

\subsection{Baselines and models}

We test different sequence labeling parsers to determine whether there are any benefits in learning across representations. We compare: (i) a single-task model for constituency parsing and another one for dependency parsing, (ii) a multi-task model for constituency parsing (and another  for dependency parsing) where each element of the 3-tuple is predicted as a partial label in a separate subtask instead of as a whole, (iii) different \textsc{mtl} models where the partial labels from a specific parsing abstraction are used as auxiliary tasks for the other one, and (iv) an \textsc{mtl} model that learns to produce both abstractions as main tasks.

\paragraph{Single-paradigm, single-task models (\single)} For constituency parsing, we use the single-task model by \citet{GomVilEMNLP2018}. The input is the raw sentence and the output for each token a single label of the form $ l_i$=$(n_i,c_i,u_i)$. For dependency parsing we use the model by \citet{viable} to predict a single dependency label of the form $r_i$=$(o_i,p_i,d_i)$ for each token.

\paragraph{Single-paradigm, multi-task models (\twoTask)} 
For constituency parsing,
instead of predicting a single label output of the form $(n_i,c_i,u_i)$, we generate three partial and separate labels $n_i$, $c_i$ and $u_i$ through three task-dependent feed-forward networks on the top of the stacked \textsc{bilstm}s. This is similar to \citet{VilaresMTL2019}.
For dependency parsing, we propose in this work a \textsc{mtl} version too.
We observed in preliminary experiments, as shown in Table \ref{tab:s-mtl}, that casting the problem as 3-task learning led to 
worse results. Instead, we cast it as a 2-task learning problem, where the first task consists in predicting the head of a word $w_i$, i.e. predicting the tuple $(o_i,p_i)$, and the second task predicts the type of the relation $(d_i)$. The loss is here computed as $\mathcal{L}$=$\sum_{t}\mathcal{L}_{t}$, where $\mathcal{L}_{t}$ is the partial loss coming from the subtask $t$.

\begin{table}[]
\centering
\scalebox{0.7}{
\begin{adjustbox}{max width=\columnwidth}
\begin{tabular}{@{}lccc@{}}
\toprule
Model & UAS & LAS \\ \midrule
\textsc{s-s} & 93.81 & 91.59 \\ \midrule
\textsc{s-mtl(2)} & \textbf{94.03} & \textbf{91.78}\\
\textsc{s-mtl(3)} & 93.66 & 91.47 \\ \bottomrule
\end{tabular}
\end{adjustbox}
}
\caption{Comparison of the single-paradigm models for dependency parsing evaluated on the PTB dev set where each label is learned as single, 2- or 3-tasks.}
\label{tab:s-mtl}
\end{table}

\begin{figure}
    \centering
    \begin{adjustbox}{max width=\columnwidth}
    \includegraphics[scale=0.40]{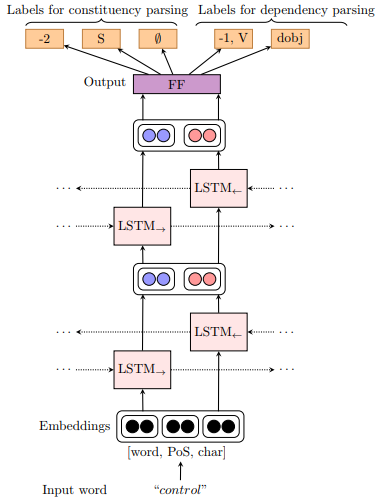}
    \end{adjustbox}
    \caption{Architecture of our double-paradigm, \textsc{mtl} model with 3-task learning for constituency parsing and 2-task learning for dependency parsing.}
    \label{fig:architecture}
\end{figure}

\paragraph{Double-paradigm, multi-task models with auxiliary losses (\aux)} We predict the partial labels from one of the parsing abstractions as main tasks. The partial labels from the other parsing paradigm are used as auxiliary tasks. The loss is computed as $\mathcal{L}$=$\sum_{t}\mathcal{L}_{t} + \sum_{a}\beta_{a}\mathcal{L}_{a}$, where $\mathcal{L}_{a}$ is an auxiliary loss and $\beta_a$ its specific weighting factor.
Figure \ref{fig:architecture} shows the architecture used in this and the following multi-paradigm model.

\paragraph{Double paradigm, multi-task models (\multi)} All tasks are learned as main tasks instead.

\begin{table}[tbp]
\centering
\begin{adjustbox}{max width=\columnwidth}
\renewcommand{\arraystretch}{.8}
\begin{tabular}{@{}cccccc@{}}
\toprule
 & \multirow{2}{*}{Model} & \multicolumn{2}{c}{\makecell{Dependency\\ Parsing}}  & \multicolumn{2}{c}{\makecell{Constituency\\ Parsing}} \\ 
\multicolumn{1}{c}{} &  & \textsc{UAS} & \textsc{LAS} &  & \textsc{F1} \\\toprule
\multirow{4}{*}{English (PTB)} & \single  & 93.60 & 91.74 &  & 90.82 \\\cmidrule(l){2-6} \cmidrule(l){2-6} 
 & \twoTask & 93.84 & 91.83&  &90.99  \\
 & \aux& \textbf{94.05}& \textbf{92.01} &  &\textbf{91.04}  \\
 & \multi &  93.96  & 91.90  &  & 90.51 \\ \midrule
 \multirow{4}{*}{Basque} & \single & 86.20 & 81.70 &  & 89.20 \\\cmidrule(l){2-6} \cmidrule(l){2-6} 
 & \twoTask &  \textbf{87.42}  & 81.71  & & 90.54  \\
& \aux & 87.19  &81.73  &  & \textbf{90.84} \\
 & \multi &87.09 & \textbf{81.77} &  &90.50  \\ \midrule
 \multirow{4}{*}{French} & \single & 89.13 & 85.03 &  & 79.71 \\\cmidrule(l){2-6} \cmidrule(l){2-6} 
 & \twoTask & \textbf{89.54} & 84.89&  & \textbf{80.40} \\
 & \aux &  89.52 & 84.97  &  &80.39 \\ 
 & \multi & 89.45 & \textbf{85.07} &  &80.24 \\ \midrule
 \multirow{4}{*}{German} & \single & 91.24 & 88.79 &  & 82.52 \\\cmidrule(l){2-6} \cmidrule(l){2-6} 
 & \twoTask & 91.54 & 88.75 & & \textbf{83.08}\\
 & \aux &  \textbf{91.58} & \textbf{88.80} &  & 82.97\\
 & \multi & 91.45 & 88.67 &  &82.87 \\ \midrule
 \multirow{4}{*}{Hebrew} & \single &  82.74 & 75.08 &  & 88.68 \\\cmidrule(l){2-6} \cmidrule(l){2-6} 
 & \twoTask &  83.42 & 74.91   &  &\textbf{91.84} \\
 & \aux & \textbf{83.90}  & \textbf{75.89}  &  & 91.78 \\
 & \multi &82.60 & 73.73 & & 91.08 \\ \midrule
 \multirow{4}{*}{Hungarian} & \single & 88.24 & 84.54 &  & 90.10 \\\cmidrule(l){2-6} \cmidrule(l){2-6} 
 & \twoTask &  88.69 & 84.54 &  & 90.51 \\
 & \aux & \textbf{88.99} & \textbf{84.95} &  &90.44  \\
 & \multi & 88.89 & 84.89 &  &\textbf{90.72}  \\ \midrule
 \multirow{4}{*}{Korean} & \single&  86.47 & 84.12 &  & 82.63 \\\cmidrule(l){2-6} \cmidrule(l){2-6} 
 & \twoTask &  86.78 & 84.39  & & \textbf{82.86} \\
 &\aux & \textbf{87.00} & \textbf{84.60} &  & 82.76 \\
 & \multi & 86.64 & 84.34& & 82.40 \\ \midrule
 \multirow{4}{*}{Polish} & \single & 91.17 & 85.64 &  & 92.59 \\\cmidrule(l){2-6} \cmidrule(l){2-6} 
 & \twoTask &  91.58 & 85.04  & &93.17 \\
& \aux & 91.37 & 85.20 &  & 93.36 \\ 
 & \multi & \textbf{92.00} &\textbf{85.92} &  & \textbf{93.52}  \\ \midrule
 \multirow{4}{*}{Swedish} & \single &86.49 & 80.60 &  & 82.56 \\\cmidrule(l){2-6} \cmidrule(l){2-6} 
 & \twoTask & 87.22 & 80.61  & & 85.16 \\
 & \aux & \textbf{87.24} & 80.34 &  &\textbf{85.49}  \\
 & \multi & 87.15 & \textbf{80.71} &  &85.38 \\ \midrule \midrule
 \multirow{4}{*}{\textit{average}} & \single &88.36 & 84.13 &  & 86.53 \\\cmidrule(l){2-6} \cmidrule(l){2-6} 
 & \twoTask & 88.89 & 84.07  & & 87.62 \\
 & \aux& \textbf{88.98}& \textbf{84.28} &  &\textbf{87.67}  \\
 & \multi & 88.80 & 84.11 &  & 87.47 \\ \midrule
\end{tabular}
\end{adjustbox}
\caption{Results on the \textsc{ptb} and \textsc{spmrl} test sets. 
}
\label{tab:main}
\end{table}

\begin{table}[tbp]
\centering
\begin{adjustbox}{max width=\columnwidth}
\begin{tabular}{@{}lccc@{}}
\toprule
\multirow{2}{*}{Model} & \multicolumn{2}{c}{\begin{tabular}[c]{@{}c@{}}Dependency \\ parsing\end{tabular}} & \begin{tabular}[c]{@{}c@{}}Constituency \\ Parsing\end{tabular} \\
 & UAS & LAS & F1 \\ \midrule
\citet{chen} & 91.80 & 89.60 &  \NA \\
\citet{kiperwasser2016simple} & 93.90 & 91.90 & \NA \\
\citet{dozat2016deep} & 95.74 & 94.08 & \NA \\
\citet{ma2018stack} & 95.87 & 94.19 & \NA \\
\defcitealias{fernandez2019left}{Fern\'andez-G and G\'omez-R (2019)}\citetalias{fernandez2019left}
& 96.04 & 94.43 & \NA \\ \midrule
\citet{vinyals2015grammar} & \NA & \NA & 88.30 \\
\citet{zhu2013fast} & \NA & \NA & 90.40 \\
\citet{VilaresMTL2019}& \NA & \NA & 91.13 \\ 
\citet{DyerRecurrent2016} & \NA &\NA  & 91.20 \\
\citet{kitaev2018constituency} &\NA &\NA & 95.13 \\  \midrule
\textbf{\textsc{d-mtl-aux}} & \textbf{94.05} & \textbf{92.01} & \textbf{91.04} \\ \bottomrule
\end{tabular}
\end{adjustbox}
\caption{Comparison of existing models against the \textsc{d-mtl-aux} model on the PTB test set.}
\label{tab:sota}
\end{table}

\begin{table*}[!tbp]
\centering
\begin{adjustbox}{max width=\textwidth}
\begin{tabular}{@{}llllllllll@{}}
\toprule
Model & Basque & French & German & Hebrew & Hungarian & Korean & Polish & Swedish & \textit{average} \\ \midrule
\citet{maltparser} & 70.11 & 77.98 & 77.81 & 69.97 & 70.15 & 82.06 & 75.63 & 73.21 & \textit{74.62} \\
\citet{ballesteros2013effective} & 78.58 & 79.00 & 82.75 & 73.01 & 79.63 & 82.65 & 79.89 & 75.82 & \textit{78.92} \\
\citet{ballesteros2015improved} \texttt{(char+POS)} & 78.61 & 81.08 & 84.49 & 72.26 & 76.34 & 86.21 & 78.24 & 74.47 & \textit{78.96} \\
\citet{de2013exploring} & 77.55 & 82.06 & 84.80 & 73.63 & 75.58 & 81.02 & 82.56 & 77.54 & \textit{79.34} \\
\citet{bjorkelund2013re} \texttt{(ensemble)} & 85.14 & 85.24 & 89.65 & 80.89 & 86.13 & 86.62 & 87.07 & 82.13 & \textit{85.36} \\ \midrule
\textbf{\textsc{D-MTL-AUX}} & 84.02 & 83.85 & 88.18 & 74.94 & 80.26 & 85.93 & 85.86 & 79.77 & \textit{82.85} \\ \bottomrule
\end{tabular}
\end{adjustbox}
\caption{Dependency parsing: existing models evaluated with \textsc{LAS} scores on the \textsc{SPMRL} test set.}
\label{tab:dependency-spmrl}
\end{table*}

\begin{table*}[tbp]
\centering
\begin{adjustbox}{max width=\textwidth}
\begin{tabular}{@{}llllllllll@{}}
\toprule
Model & Basque & French & German & Hebrew & Hungarian & Korean & Polish & Swedish & \textit{average} \\ \midrule
\citet{fernandez-gonzalez-martins-2015-parsing} & 85.90 & 78.75 & 78.66 & 88.97 & 88.16 & 79.28 & 91.20 & 82.80 & \textit{84.22} \\
\citet{coavoux2016neural} & 86.24 & 79.91 & 80.15 & 88.69 & 90.51 & 85.10 & 92.96 & 81.74 & \textit{85.66} \\
\citet{bjorkelund2013re} \texttt{(ensemble)}  & 87.86 &81.83  &81.27  &89.46  &91.85  &84.27  &87.55  &83.99  & \textit{86.01} \\
\citet{VilaresMTL2019} & 90.85 & 80.40 & 83.42 & 92.05 & 90.38 & 83.24 & 93.93 & 85.54 & \textit{87.48} \\
\citet{aux} & 88.81 & 82.49 & 85.34 & 89.87 & 92.34 & 86.04 & 93.64 & 84.00 & \textit{87.82} \\
\citet{kitaev2018constituency}& 89.71 & 84.06 & 87.69 & 90.35 & 92.69 & 86.59 & 93.69 & 84.35 & \textit{88.64} \\ \midrule
\textbf{\textsc{D-MTL-AUX}} & 90.84 & 80.39 & 82.97 & 91.78 & 90.44 & 82.76 & 93.36 & 85.49 & \textit{87.25} \\ \bottomrule
\end{tabular}
\end{adjustbox}
\caption{Constituency parsing: existing models evaluated with \textsc{F1} score on the \textsc{SPMRL} test set.}
\label{tab:constituency-spmrl}
\end{table*}

\begin{table}[tbp]
\centering
\begin{adjustbox}{max width=\columnwidth}
\renewcommand{\arraystretch}{.8}
\begin{tabular}{@{}ccc@{}}
\toprule
Model  & Dependency parsing & Constituency parsing 
 \\\midrule
\single & \stddev{102}{6} & \stddev{117}{6}   \\
\twoTask &  \stddev{128}{11} &  \stddev{133}{1}  \\
 \aux&  \stddev{128}{11} &  \stddev{133}{1}    \\
 \multi&  \stddev{124}{1} &  \stddev{124}{1}   \\
 \bottomrule
\end{tabular}
\end{adjustbox}
\caption{Sentences/second on the \textsc{ptb} test set.  }
\label{tab:speed}
\end{table}

\section{Experiments}

\subsection{Data} 
In the following experiments we use two parallel datasets that provide syntactic analyses for both dependency and constituency parsing.
\paragraph{PTB} For the evaluation on English language we use the English Penn Treebank \cite{marcus1993building},
transformed
into Stanford dependencies \cite{deMarneffe} with the predicted PoS tags as in \citet{DyerRecurrent2016}. 
\paragraph{SPMRL}
We also use the \textsc{spmrl} datasets, a collection of parallel dependency and constituency treebanks for morphologically rich languages \cite{seddah2014introducing}. In this case, we use the predicted PoS tags provided by the organizers. We observed some differences between the constituency and dependency predicted input features provided with the corpora. For experiments where dependency parsing is the main task, we use the input from the dependency file, and 
the converse for constituency, for comparability with other work.
\textsc{d-mtl} models were trained twice
(one for each input), and dependency and constituent scores are reported on the model trained on the corresponding input.

\paragraph{Metrics} We use bracketing F-score from the original \textsc{evalb} (together with COLLINS.prm) and \textsc{eval\_spmrl} (together with spmrl.prm) official scripts to evaluate constituency trees. For dependency parsing, we rely on \textsc{las} and \textsc{uas} scores where punctuation is excluded in order to provide a homogeneous setup for \textsc{PTB} and \textsc{SPMRL}.

\subsection{Results}

Table \ref{tab:main} compares
single-paradigm models
against their double-paradigm \textsc{mtl} versions.
On average, 
\textsc{mtl} models with auxiliary losses 
achieve the best performance for both parsing abstractions. They gain $1.14$
\textsc{F1} points on average
in comparison with the single model for constituency parsing, and $0.62$ \textsc{uas} and $0.15$ \textsc{las} points for dependency parsing. In comparison to the single-paradigm MTL models, the average gain is smaller: 0.05 \fone{} points for constituency parsing, and 0.09 \uas{} and 0.21 \las{} points for dependency parsing.

\textsc{mtl} models that use auxiliary tasks (\aux{}) consistently outperform the single-task models (\single{}) in all datasets, both for constituency parsing and for dependency parsing in terms of \textsc{uas}. However, this does not extend to \textsc{las}.
This different behavior between \uas{} and \las{} seems to be originated by the fact that 2-task dependency parsing models, which are the basis for the corresponding auxiliary task and \mtl{} models, improve \uas{} but not \las{} with respect to single-task dependency parsing models. The reason might be that the single-task setup excludes unlikely
combinations of dependency labels with PoS tags or dependency directions
that are not found in the training set, while in the 2-task setup, both components are treated separately, which may be having a negative influence on dependency labeling accuracy. 

In general, one can observe different range of gains of the models across languages. In terms of \textsc{uas}, the 
differences between single-task and \mtl{} models
span between $1.22$ (Basque) and $-0.14$ (Hebrew); for \textsc{las}, $0.81$ and $-1.35$ (both for Hebrew); and for \textsc{F1}, $3.16$ 
(Hebrew) and $-0.31$ (English). 
Since 
the sequence labeling encoding used
for dependency parsing heavily relies on PoS tags, the result for a given language can be dependent on the degree of the granularity of its PoS tags.

In addition, Table \ref{tab:sota} provides a comparison of the \textsc{d-mtl-aux} models for dependency and constituency parsing against  existing models 
on the PTB test set. Tables \ref{tab:dependency-spmrl} and \ref{tab:constituency-spmrl} shows the results for various existing models on the \textsc{SPMRL} test sets.\footnote{Note that we provide these SPMRL results for merely informative purposes. While they are the best existing results to our knowledge in these datasets, not all are directly comparable to ours (due to not all of them using the same kinds of information, e.g. some models do not use morphological features). Also, there are not many recent results for dependency parsing on the \textsc{SPMRL} datasets, probably due to the popularity of UD corpora. For comparison, we have included punctuation for the dependency parsing evaluation. 
}

Table \ref{tab:speed} shows the speeds (sentences/second) on a single core of a CPU\footnote{Intel Core i7-7700 CPU 4.2 GHz.}. The \textsc{d-mtl} setup comes at almost no added computational cost, so the very good speed-accuracy tradeoff already provided by the single-task models is improved.

\section{Conclusion}
We have
described a framework to  leverage
the complementary nature of constituency and dependency parsing. 
It combines multi-task learning, auxiliary tasks, and 
sequence labeling parsing, 
so that
constituency and dependency parsing can benefit each other through learning across their 
representations. 
We have shown that \textsc{mtl} models with auxiliary losses outperform single-task models,
and
\textsc{mtl} models that treat both constituency and dependency parsing as main tasks 
obtain strong results, 
coming
almost at no cost in terms of speed.
Source code will be released upon acceptance.

\section*{Acknowlegments}

This work has received funding from the
European Research Council (ERC), under the European Union's Horizon 2020 research and innovation programme (FASTPARSE, grant agreement No 714150), from the 
ANSWER-ASAP project (TIN2017-85160-C2-1-R) from MINECO, and from Xunta de Galicia (ED431B 2017/01).

\bibliography{acl2019}
\bibliographystyle{acl_natbib}

\clearpage
\appendix

\section{Model parameters}\label{appendix-training-configuration}

The models were trained up to 150 iterations and optimized with Stochastic Gradient Descent (SGD) with a batch size of 8. The best model for constituency parsing was chosen with the highest achieved \textsc{F1} score on the development set during the training and for dependency parsing with the highest \textsc{las} score. The best double paradigm, multi-task model was chosen based on the highest harmonic mean among \textsc{las} and \textsc{F1} scores. 

Table~\ref{tab:hyper} shows model hyperparameters.

\begin{table}[hbtp]
\centering
\begin{adjustbox}{max width=\columnwidth}
\renewcommand{\arraystretch}{0.9}
\begin{tabular}{ll}
\hline
Initial learning rate & 0.02 \\
Time-based learning rate decay & 0.05 \\
Momentum & 0.9 \\
Dropout & 0.5 \\
\midrule 
&Dimension\\ \midrule
Word embedding & 100 \\
Char embedding & 30 \\
Self-defined features & 20 \tablefootnote{Models trained on PTB treebank used PoS tag embedding size of 25 in order to assure the same setup for comparison with the previously reported results.} \\
Word hidden vector & 800 \\
Character hidden vector & 50 \\
\midrule 
Type of \textsc{mtl} model& Weighting factor \\&for each task\\ \midrule
2-task $D$& 1  \\
3-task $C$ & 1 \\
$D$ with auxiliary task $C$  & $D$: 1 and $C$: 0.2 \\
$C$ with auxiliary task $D$  & $C$: 1 and $D$: 0.1 \\
Multi-task $C$ and $D$ & 1 \\
 \hline
\end{tabular}
\end{adjustbox}
\caption{Model hyperparameters. $D$ indicates dependency parsing and $C$ constituency parsing. }
\label{tab:hyper}
\end{table}

\end{document}